\documentclass[]{spie} 

\usepackage{amssymb}
\usepackage{mwe} 
\usepackage{bm}
\usepackage{subcaption}
\usepackage{multirow}
\usepackage{amsmath,scalerel}
\usepackage{xcolor}

\makeatletter
\newcommand*\bigcdot{\mathpalette\bigcdot@{.5}}
\newcommand*\bigcdot@[2]{\mathbin{\vcenter{\hbox{\scalebox{#2}{$\m@th#1\bullet$}}}}}

\title{Weakly-supervised Mamba-Based Mastoidectomy Shape Prediction for Cochlear Implant Surgery Using 3D T-Distribution Loss}

\author[a]{Yike Zhang}
\author[a,b]{Jack H. Noble}
\affil[a]{Dept. of Computer Science, Vanderbilt University}
\affil[b]{Dept. of Electrical and Computer Engineering, Vanderbilt University}

\begin{document}
\maketitle
\begin{abstract}
Cochlear implant surgery is a treatment for individuals with severe hearing loss. It involves inserting an array of electrodes inside the cochlea to electrically stimulate the auditory nerve and restore hearing sensation. A crucial step in this procedure is mastoidectomy, a surgical intervention that removes part of the mastoid region of the temporal bone, providing a critical pathway to the cochlea for electrode placement. Accurate prediction of the mastoidectomy region from preoperative imaging assists presurgical planning, reduces surgical risks, and improves surgical outcomes. In previous work, a self-supervised network was introduced to predict the mastoidectomy region using only preoperative CT scans. While promising, the method suffered from suboptimal robustness, limiting its practical application. To address this limitation, we propose a novel weakly-supervised Mamba-based framework to predict accurate mastoidectomy regions directly from preoperative CT scans. Our approach utilizes a 3D T-Distribution loss function inspired by the Student-\textit{t} distribution, which effectively handles the complex geometric variability inherent in mastoidectomy shapes. Weak supervision is achieved using the segmentation results from the prior self-supervised network to eliminate the need for manual data cleaning or labeling throughout the training process. The proposed method is extensively evaluated against state-of-the-art approaches, demonstrating superior performance in predicting accurate and clinically relevant mastoidectomy regions. Our findings highlight the robustness and efficiency of the weakly-supervised learning framework with the proposed novel 3D T-Distribution loss. To the best of our knowledge, this is the first work to employ weakly-supervised learning for mastoidectomy shape prediction, achieving a higher benchmark Dice score of 0.72 compared with prior baseline in this critical application area. These advancements not only pave the way for improved cochlear implant surgical preoperative planning but also contribute to the broader adoption of using 3D T-distribution loss in weakly-supervised techniques among 3D medical imaging-related tasks.
\end{abstract}

\section{Introduction}
\label{sec:intro}
\begin{figure}[ht]
  \centering
  \begin{minipage}{0.115\textwidth}
        \includegraphics[width=\textwidth]{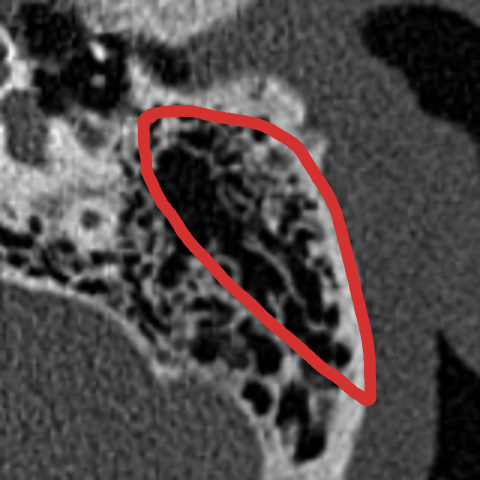}
    \end{minipage}
    \vspace{0.4em}
    \begin{minipage}{0.115\textwidth}
        \includegraphics[width=\textwidth]{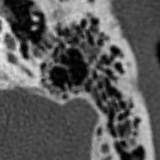}
    \end{minipage}
    \begin{minipage}{0.115\textwidth}
        \includegraphics[width=\textwidth]{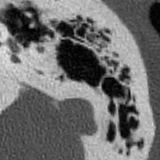}
    \end{minipage}
    \begin{minipage}{0.115\textwidth}
        \includegraphics[width=\textwidth]{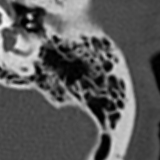}
    \end{minipage}
    \begin{minipage}{0.115\textwidth}
        \includegraphics[width=\textwidth]{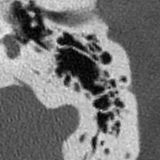}
    \end{minipage}
    \begin{minipage}{0.115\textwidth}
        \includegraphics[width=\textwidth]{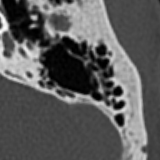}
    \end{minipage}
    \begin{minipage}{0.115\textwidth}
        \includegraphics[width=\textwidth]{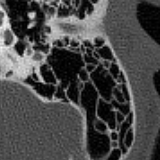}
    \end{minipage}
    \begin{minipage}{0.115\textwidth}
        \includegraphics[width=\textwidth]{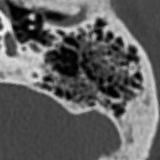}
    \end{minipage}
    \begin{minipage}{0.115\textwidth}
        \centering
        \includegraphics[width=\textwidth]{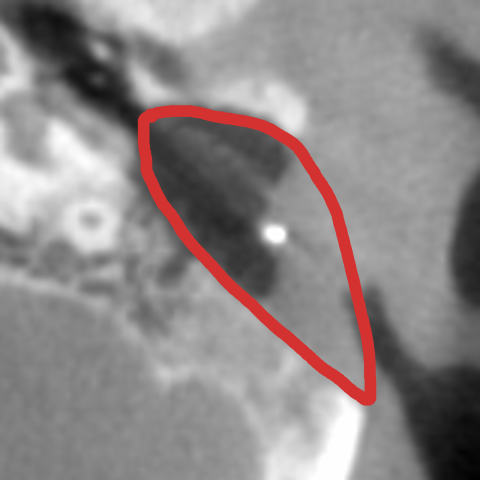}
        \footnotesize{Sample 1}
    \end{minipage}
    \begin{minipage}{0.115\textwidth}
        \centering
        \includegraphics[width=\textwidth]{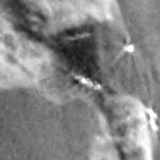}
        \footnotesize{Sample 2}
    \end{minipage}
    \begin{minipage}{0.115\textwidth}
        \centering
        \includegraphics[width=\textwidth]{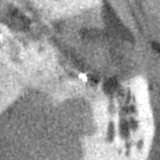}
        \footnotesize{Sample 3}
    \end{minipage}
    \begin{minipage}{0.115\textwidth}
        \centering
        \includegraphics[width=\textwidth]{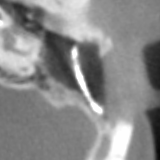}
        \footnotesize{Sample 4}
    \end{minipage}
    \begin{minipage}{0.115\textwidth}
        \centering
        \includegraphics[width=\textwidth]{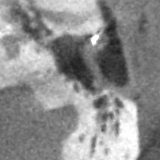}
        \footnotesize{Sample 5}
    \end{minipage}
    \begin{minipage}{0.115\textwidth}
        \centering
        \includegraphics[width=\textwidth]{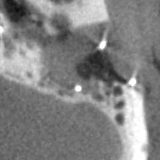}
        \footnotesize{Sample 6}
    \end{minipage}
    \begin{minipage}{0.115\textwidth}
        \centering
        \includegraphics[width=\textwidth]{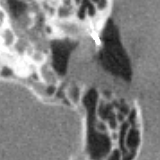}
        \footnotesize{Sample 7}
    \end{minipage}
    \begin{minipage}{0.115\textwidth}
        \centering
        \includegraphics[width=\textwidth]{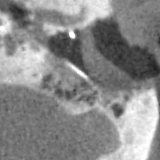}
        \footnotesize{Sample 8}
    \end{minipage}
\caption{\textbf{Example CT scans and Mastoidectomy Region}. \textbf{Top row} shows preoperative CT scans and \textbf{bottom row} illustrates the corresponding highly noisy postoperative CT scans. The highlighted area in red is an example of the mastoidectomy shape region.}
\label{fig:preop_postop}
\end{figure}

Cochlear implant surgery is performed by otologists to restore hearing in patients with moderate-to-profound hearing loss \cite{labadie2018preliminary} using an array of electrodes. Before placing the electrodes, surgeons typically perform a mastoidectomy to remove part of the mastoid portion of the temporal bone to provide safe access to the cochlear using a high-speed drill. Figure~\ref{fig:preop_postop} shows the final mastoidectomy cavity in postoperative CT scans. As seen in the figure, postoperative CT scans contain metal artifacts from implant electrode arrays and intensity heterogeneity due to fluid accumulation in the ear canal after surgery. Our research goal is to predict the mastoidectomy region directly from preoperative CT scans to assist the preoperative planning process.
However, the removed mastoid bone does not form a regular or enclosed shape, as the mastoidectomy region contains pneumatized bone with numerous small air cells. Consequently, popular segmentation methods, such as the SAM or SAM2-derived medical imaging segmentation models \cite{SAM4MIS, lou2024zeroshotsurgicaltoolsegmentation} may not be suitable for this challenging task due to the irregular and boundary-ambiguous shape of the mastoid region \cite{kirillov2023segment, ravi2024sam2segmentimages}. 
Another potential approach involves image synthesis to predict the mastoidectomy region in preoperative scans by using them to synthesize postoperative images. However, due to the noise and artifacts present in postoperative images, synthesizing networks are likely to generate undesirable artifacts, such as false electrodes or wiring, making this approach impractical for our research goal. 
Previous studies have explored alternative methods for predicting the shape of the mastoidectomy. For example, \cite{Dillon2015A} utilized a bone-attached robotic system to perform the mastoidectomy automatically. While their method demonstrated accuracy in performing the procedures, it requires surgeons to manually label the mastoidectomy region on preoperative CT scans, which is time-consuming and inefficient. Most recently, \cite{zhang2024mmunsupervisedmambabasedmastoidectomy} proposed a self-supervised neural network for automatic prediction of the mastoidectomy shape region from any given preoperative CT scan, achieving a Dice score of 0.702 between the predicted region and ground truth labels.
In this study, we aim to build on these efforts by utilizing preoperative CT scans as inputs to a neural network, along with the weak labels that are generated from method \cite{zhang2024mmunsupervisedmambabasedmastoidectomy}, to improve the mastoidectomy shape prediction results. We propose a weakly-supervised method that uses the 3D T-Distribution loss to enhance the prediction accuracy of mastoidectomy shapes. Our results demonstrate that 3D T-Distribution loss is a robust loss function for weakly-supervised tasks, particularly in 3D medical imaging where weak labels are often used. By combining weakly-supervised learning with the proposed 3D T-Distribution loss, the need for labor-intensive manual labeling is eliminated, making the approach generalizable for a wide range of medical segmentation tasks, even with uncleaned and highly noisy labels.
\section{Method}
\label{sec:method}
\subsection{Proposed Framework}
\begin{figure}[htb]
\centering
\begin{minipage}[b]{\linewidth}
  \includegraphics[width=\textwidth]{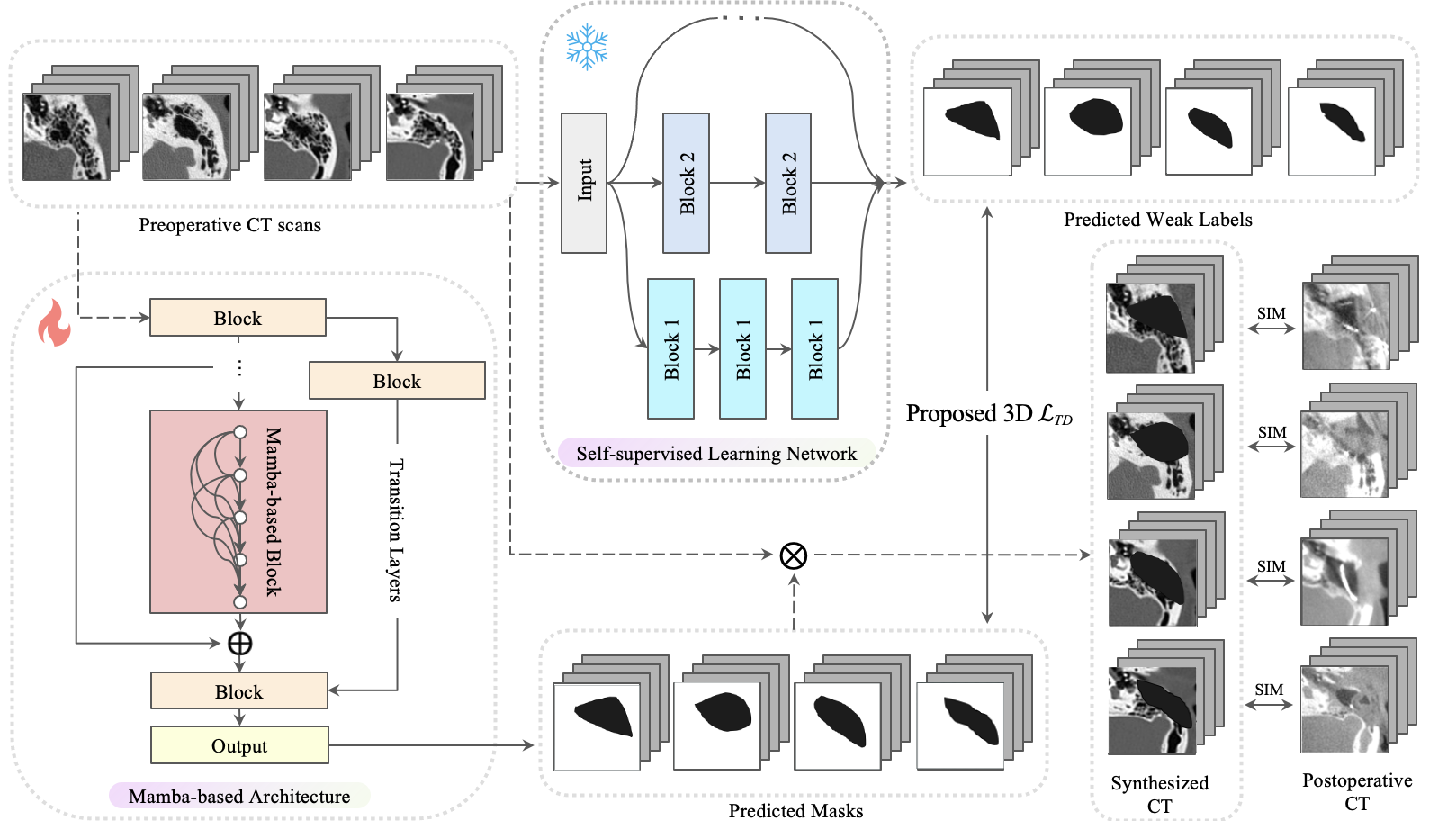}
  \caption{\textbf{Framework Overview}. \textcolor{cyan}{Snowflake} indicate frozen and non-trainable parameters, \textcolor{red}{fire} represents trainable parameters, and ``SIM" compares the similarity between two objects.}
\label{fig:overview}
\end{minipage}
\end{figure}

Figure~\ref{fig:overview} demonstrates the training and inference pipeline in our proposed framework. The preoperative scans are represented as $\bm{I} \in \mathbb{R}^{D \times H \times W}$, where $D$, $H$, and $W$ correspond to the depth, height, and width dimensions of the CT scans. In our dataset, $D=64$, $H=160$, and $W=160$. These volumetric images are then input into the SegMamba-based neural network $f_{\bm{\theta}}(\bm{I})$, which outputs probability masks $\bm{Y} \in \mathbb{R}^{D \times H \times W}$, representing the predicted mastoidectomy shape. During the weakly-supervised training process, we refine these predictions by comparing $\bm{Y}$ with weak labels $\bm{K} \in \{0, 1\}^{D \times H \times W}$, which are generated according to the method proposed in \cite{zhang2024mmunsupervisedmambabasedmastoidectomy}. To further improve the accuracy of the predicted masks and mitigate the noise and large errors introduced by the weak labels, we propose the 3D T-Distribution loss, $\mathcal{L}_{TD}(\bm{Y}, \bm{K})$, which refines the probability masks and aligns them more closely with the actual mastoidectomy shape region.
\subsection{Network Design}
\begin{figure}[htb]
\begin{minipage}[b]{1.0\linewidth}
  \centering
  \centerline{\includegraphics[width=\textwidth]{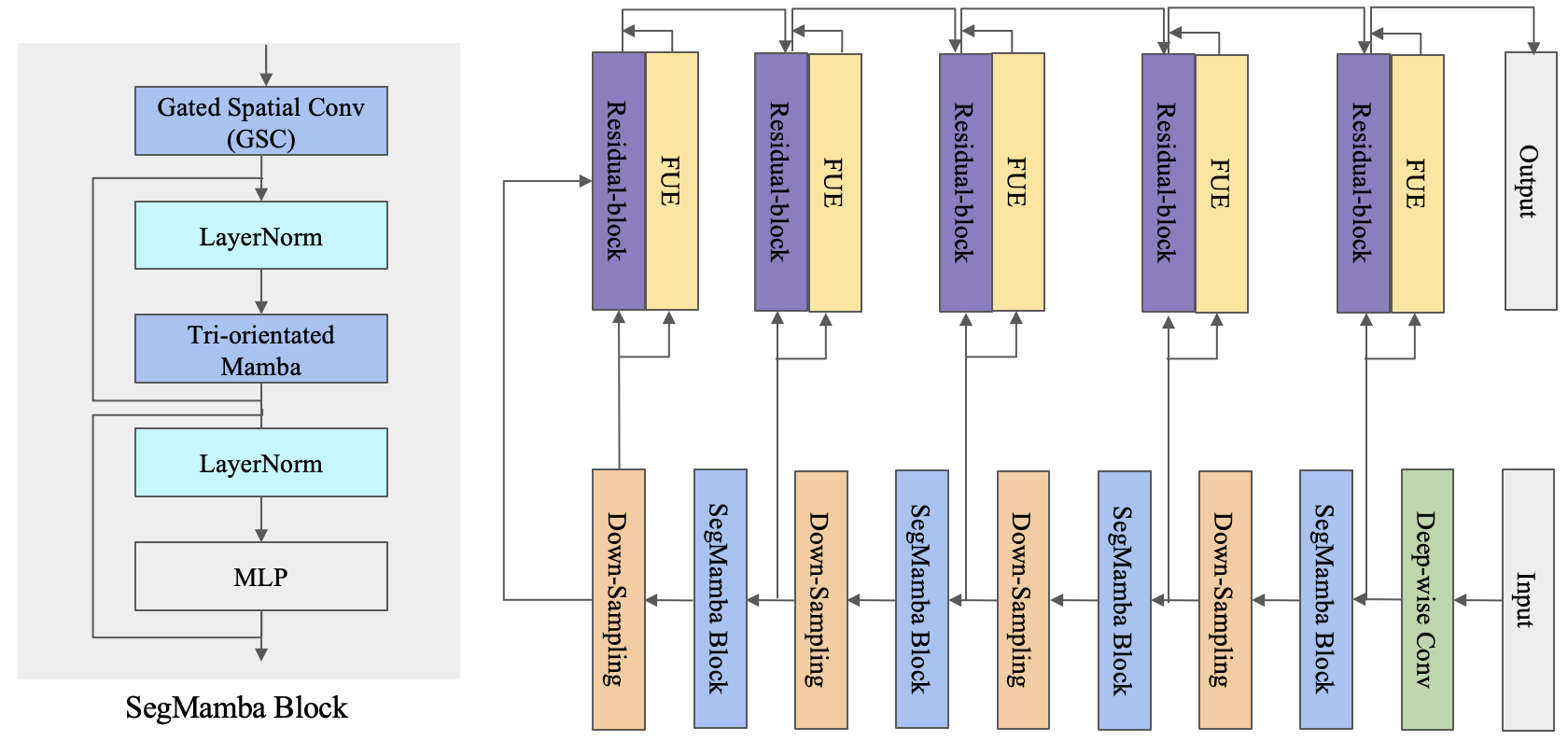}}
\caption{\textbf{Mamba-based Architecture}. The model takes preoperative CT as input and outputs the predicted mastoidectomy shape.}
\label{fig:segmamba}
\end{minipage}
\end{figure}

In this study, we use the state-of-the-art SegMamba \cite{xing2024segmambalongrangesequentialmodeling} neural network architecture shown in Figure~\ref{fig:segmamba} to train the mastoidectomy shape prediction network $f_{\bm{\theta}}$. This is a novel framework that combines the U-Net \cite{ronneberger2015unetconvolutionalnetworksbiomedical} shape-like structure with the newly released Mamba \cite{gu2024mambalineartimesequencemodeling} for modeling the global features in 3D image volumes at various scales. 
Mamba is developed based on state space models and aims to capture long-range dependencies and enhance the efficiency of training and inference. SegMamba architecture contains a gated spatial convolution (GSC) module, the tri-orientated Mamba (ToM) module, and a feature-level uncertainty estimation (FUE) module. GSC module improves the spatial feature representation before each ToM module, ToM module supports whole-volume sequential modeling of 3D features, and FUE module selects and reuses the multi-scale features from the encoder.
\subsection{3D T-Distribution Loss}
The 3D T-Distribution loss function extends the 2D-imaging-focused method proposed by \cite{gonzalezjimenezRobustTLoss2023} and is specifically tailored for 3D imaging segmentation tasks. T-Distribution loss is derived from the negative log-likelihood of the Student-\textit{t} distribution, which is known for its ability to model data with heavy tails. The Probability Density Function (PDF) of the multivariate continuous Student-\textit{t} distribution is defined in Eq~\ref{eq:pdf}. 
\begin{equation}
    P(\bm{K}|\bm{\mu},\bm{\Sigma};r)=
    (\pi r)^{-D'/2} \; \bigcdot \frac{\Gamma(\frac{r+D'}{2})}{\Gamma(\frac{r}{2})} \; \bigcdot |\det(\bm{\Sigma})|^{-1/2} \; \bigcdot \\
    \left[ 1 + \frac{(\bm{K}-\bm{\mu})^T \bm{\Sigma}^{-1}(\bm{K}-\bm{\mu})}{r} \right]^{-\frac{r+D'}{2}},
\label{eq:pdf}
\end{equation} where $r$ represents the distribution's degree of freedom, $D'$ is the dimensional variable, $\bm{\mu}$ is the mean, and $\bm{\Sigma}$ is the covariance matrix of the distribution. Different from the Gaussian distribution, which is sensitive to outliers, the Student-\textit{t} distribution has a broader bell shape, making it more robust when encountering data points that deviate significantly from the norm. The robustness in T-Distribution loss is achieved by controlling the variable $r$. A smaller $r$ creates a heavier tail thus reducing the influence of outliers by giving less weight to large errors. In contrast, a larger $r$ makes the distribution more Gaussian, increasing the sensitivity to outliers. Given this attribute, T-Distribution loss can be tuned to work with imperfect datasets more effectively than traditional loss functions. This makes it particularly valuable in medical image processing, where outliers and noise are common due to artifacts or variations in imaging quality. Furthermore, when applying T-Distribution loss to the weakly-supervised learning task where noisy labels are used during training, it helps improve the accuracy of predictions by mitigating the impact of outliers, leading to robust and reliable outcomes. The negative log-likelihood loss term can be represented in Eq~\ref{Eq: tloss}:

\begin{multline}
        \mathcal{L}_{TD} = \frac{D'}{2}\log(\pi r) + \log \Gamma \left(\frac{r}{2}\right) - \log \Gamma \left(\frac{r + D'}{2}\right) + \frac{1}{2}\sum_{i=1}^{D'}\log\sigma_{i}^{2} + \frac{r+D'}{2} \log \left[1 + \frac{1}{r}\sum_{i=1}^{D'}\frac{(K_i - u_i)^2}{\sigma_{i}^{2}}\right]
\label{Eq: tloss}
\end{multline} We set $\delta$ = $\bm{K} - f_{\bm{\theta}}(\bm{I})$ and $\sigma^{2}$ are the diagonal elements of the covariance matrix $\bm{\Sigma}$. Since $\bm{\Sigma}$ is symmetric and positive semi-definite, it is possible to only calculate the diagonal and lower triangular elements because the upper triangular part is simply mirroring of the lower triangular part. However, the dimensional data in our dataset is large, thus optimizing the full covariance matrix becomes computationally expensive and it is not feasible to be implemented under our current experiment settings.
In our experiment, we improve the overall performance by configuring the covariance matrix, $\bm{\Sigma}$, as a diagonal matrix rather than an identity matrix. By updating the parameter $r$ and $\sigma^{2}$ alongside the network parameters $\bm{\theta}$ during backpropagation, the network dynamically adapts to varying levels and distributions of weak labels without prior knowledge. Since the network's predictions may include negative values, we apply the SoftPlus activation function to $\log\sigma_{i}^{2}$ to ensure the output values remain positive.
\section{Experiments and Results}
\label{sec:results}
\subsection{Implementation Details}
To increase variations in the training data, we perform the following data augmentation techniques: random swapping, flipping, elastic deformations, and affine transformations. The initial learning rates for $\theta$, $r$, and $\sigma^2$, are set to $1 \times 10^{-3}$, $1 \times 10^{-4}$, and $1 \times 10^{-4}$, respectively. The 3D T-Distribution loss is initialized with $r$ = 1 and $\sigma^{2}$ as an identity matrix. A safeguard number $1e-8$ is applied to both of them for numerical stability. All models are trained for up to 600 epochs on an NVIDIA RTX 4090 GPU with an early stopping technique based on $\mathcal{L}_{TD}$.
\subsection{Evaluation Metrics}
The Dice similarity coefficient (Dice) and Hausdorff Distance 95\% (HD95) are used as part of the evaluation metrics for experiment results. HD95 calculates the 95th percentile of surface distances between ground truth and prediction point sets. Metric formulations are presented in the following:

\begin{equation}
    \text{Dice} = \frac{2\Sigma_{i=1}^N Y_i \bar{Y}_i}{\Sigma_{i=1}^N Y_i + \Sigma_{i=1}^N  \bar{Y_i}},
\end{equation} where N denotes the number of pixels. $Y$ and $\bar{Y}$ represent the output probability and ground truth masks, respectively.

\begin{equation}
    \text{HD95} = \max \{\underset{y^\prime \in Y^\prime}{\max} \underset{\bar{y^\prime}\in \bar{Y}^{\prime}}{\min} || y^\prime - \bar{y}^\prime ||_{0.95}, \underset{\bar{y}^\prime \in Y^\prime}{\max} \underset{y^\prime \in \bar{Y}^{\prime}}{\min} || \bar{y}^\prime - y^\prime||_{0.95} \},
\end{equation} where $y^\prime$ and $\bar{Y}^{\prime}$ denote ground truth and prediction surface point sets. Average Surface Distance (ASD) is also used in the evaluation. Unlike volume-based overlap metrics, such as Dice Score and Specificity, the HD95 and ASD evaluate the agreement between the surfaces of the ground truth labels and predicted structures, with a fixed tolerance. This offers a more precise evaluation of how closely those two surfaces align.
\begin{figure}[ht]
    \centering
    \begin{minipage}{\textwidth}
        \begin{minipage}[c]{0.00001\textwidth}
            \centering  
            \rotatebox{90}{\footnotesize Preoperative CT Scan}
        \end{minipage}
        \begin{minipage}[c]{\textwidth}
            \centering
            \begin{minipage}[b]{0.23\textwidth}
                \centering
                \includegraphics[width=\textwidth]{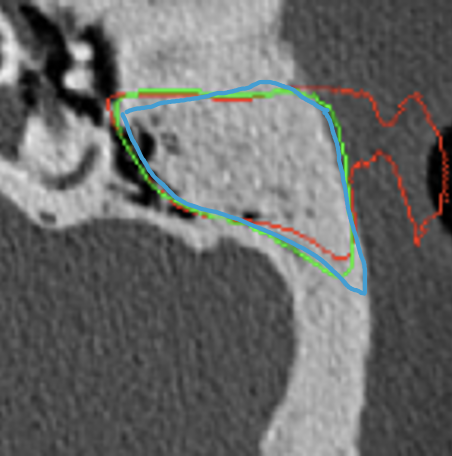}
            \end{minipage}
            \begin{minipage}[b]{0.23\textwidth}
                \centering
                \includegraphics[width=\textwidth]{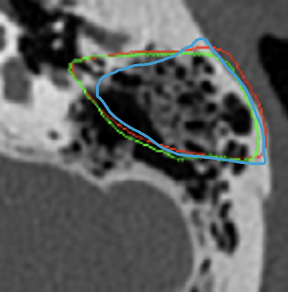}
            \end{minipage}
            \begin{minipage}[b]{0.23\textwidth}
                \centering
                \includegraphics[width=\textwidth]{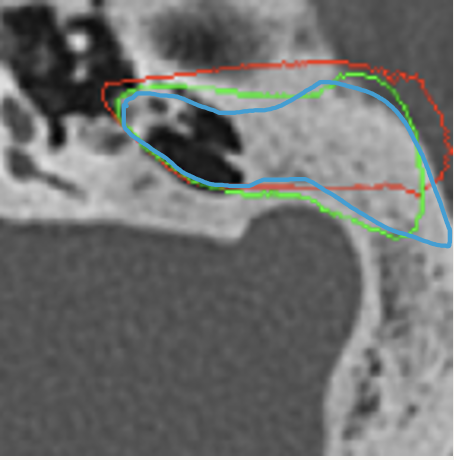}
            \end{minipage}
            \begin{minipage}[b]{0.23\textwidth}
                \centering
                \includegraphics[width=\textwidth]{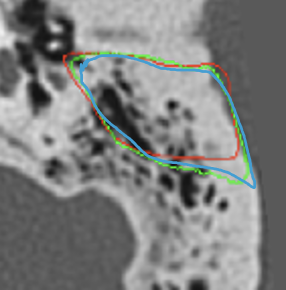}
            \end{minipage}
        \end{minipage}
    \end{minipage}
    
    \vspace{0.2em}
    
    \begin{minipage}{\textwidth}
        \begin{minipage}[c]{0.00001\textwidth}
            \centering
            \rotatebox{90}{\footnotesize Postoperative CT Scan}
        \end{minipage}
        \begin{minipage}[c]{\textwidth}
            \centering
            \begin{minipage}[b]{0.23\textwidth}
                \centering
                \includegraphics[width=\textwidth]{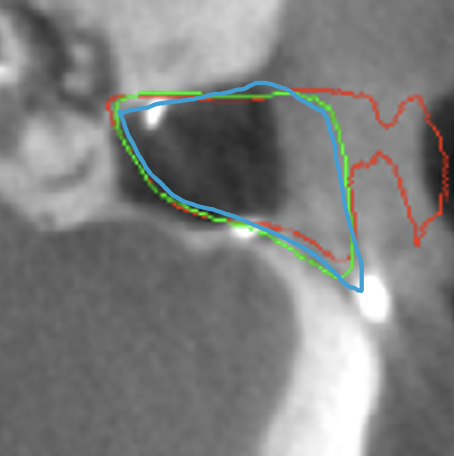}
            \end{minipage}
            \begin{minipage}[b]{0.23\textwidth}
                \centering
                \includegraphics[width=\textwidth]{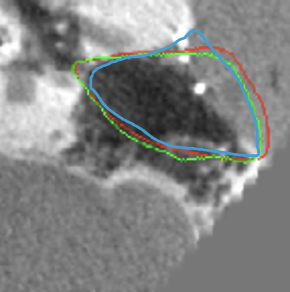}
            \end{minipage}
            \begin{minipage}[b]{0.23\textwidth}
                \centering
                \includegraphics[width=\textwidth]{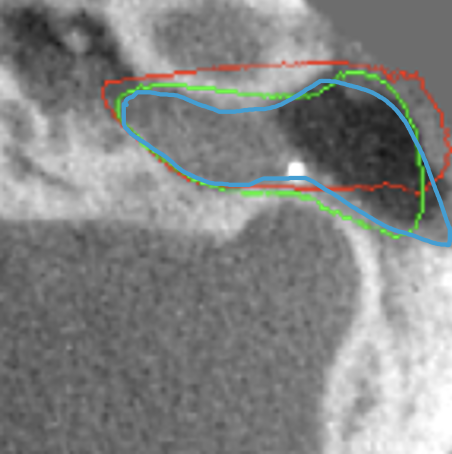}
            \end{minipage}
            \begin{minipage}[b]{0.23\textwidth}
                \centering
                \includegraphics[width=\textwidth]{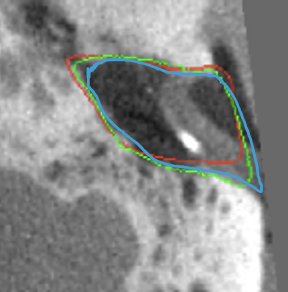}
            \end{minipage}
        \end{minipage}
    \end{minipage}
    \begin{minipage}{\textwidth}
        \hfill
        \begin{minipage}[c]{0.23\textwidth}
            \centering
            \footnotesize{Sample 1}
        \end{minipage}\hfill
        \begin{minipage}[c]{0.23\textwidth}
            \centering
            \footnotesize{Sample 2}
        \end{minipage}\hfill
        \begin{minipage}[c]{0.23\textwidth}
            \centering
            \footnotesize{Sample 3}
        \end{minipage}\hfill
        \begin{minipage}[c]{0.23\textwidth}
            \centering
            \footnotesize{Sample 4}
        \end{minipage}
        \hfill
    \end{minipage}

    \caption{\textbf{Mastoidectomy Shape Prediction}. Ground truth labels are highlighted in blue, predictions from the proposed method in green, and the baseline method from \cite{zhang2024mmunsupervisedmambabasedmastoidectomy} in red.}
    \label{fig:ws_output}
\end{figure}
\subsection{Evaluation Results}
\begin{figure}[ht]
    \centering
    \begin{minipage}[t]{0.10\textwidth}
        \centering
        \captionsetup{labelformat=empty}
        \includegraphics[width=\textwidth]{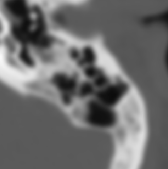}\\
        \vspace{0.4em}
        \includegraphics[width=\textwidth]{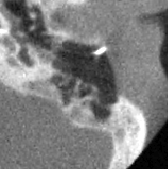}
        \footnotesize{CT Scan}
    \end{minipage}%
    \hfill
    \begin{minipage}[t]{0.10\textwidth}
        \centering
        \captionsetup{labelformat=empty}
        \includegraphics[width=\textwidth]{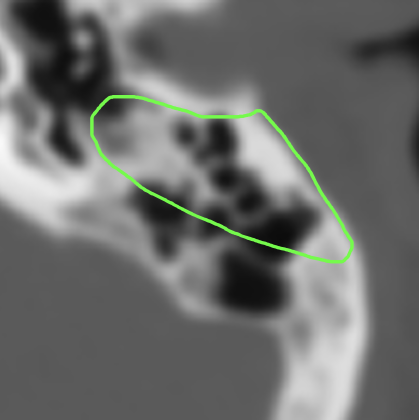}\\
        \vspace{0.4em}
        \includegraphics[width=\textwidth]{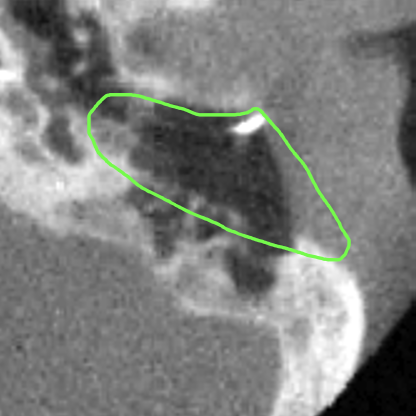}
        \footnotesize{Ground Truth}
    \end{minipage}%
    \hfill
    \begin{minipage}[t]{0.10\textwidth}
        \centering
        \captionsetup{labelformat=empty}
        \includegraphics[width=\textwidth]{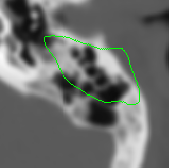}\\
        \vspace{0.4em}
        \includegraphics[width=\textwidth]{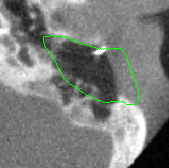}
        \footnotesize{Our Method}
    \end{minipage}%
    \hfill
    \begin{minipage}[t]{0.10\textwidth}
        \centering
        \captionsetup{labelformat=empty}
        \includegraphics[width=\textwidth]{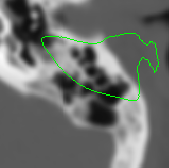}\\
        \vspace{0.4em}
        \includegraphics[width=\textwidth]{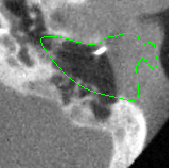}
        \footnotesize{Baseline}
    \end{minipage}%
    \hfill
    \begin{minipage}[t]{0.10\textwidth}
        \centering
        \captionsetup{labelformat=empty}
        \includegraphics[width=\textwidth]{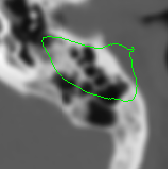}\\
        \vspace{0.4em}
        \includegraphics[width=\textwidth]{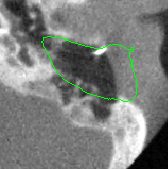}
        \footnotesize{SwinUNETR}
    \end{minipage}%
    \hfill
    \begin{minipage}[t]{0.10\textwidth}
        \centering
        \captionsetup{labelformat=empty}
        \includegraphics[width=\textwidth]{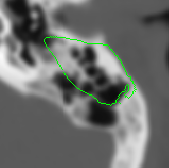}\\
        \vspace{0.4em}
        \includegraphics[width=\textwidth]{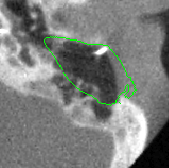}
        \footnotesize{UNETR}
    \end{minipage}%
    \hfill
    \begin{minipage}[t]{0.10\textwidth}
        \centering
        \captionsetup{labelformat=empty}
        \includegraphics[width=\textwidth]{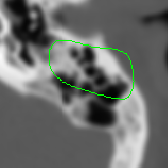}\\
        \vspace{0.4em}
        \includegraphics[width=\textwidth]{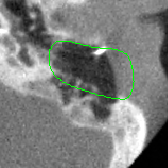}
        \footnotesize{R2AttUNET}
    \end{minipage}%
    \hfill
    \begin{minipage}[t]{0.10\textwidth}
        \centering
        \captionsetup{labelformat=empty}
        \includegraphics[width=\textwidth]{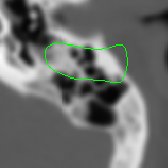}\\
        \vspace{0.4em}
        \includegraphics[width=\textwidth]{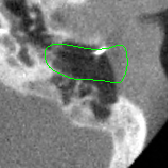}
        \footnotesize{R2UNET}
    \end{minipage}%
    \hfill
    \begin{minipage}[t]{0.10\textwidth}
        \centering
        \captionsetup{labelformat=empty}
        \includegraphics[width=\textwidth]{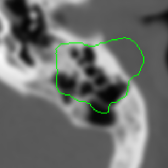}\\
        \vspace{0.4em}
        \includegraphics[width=\textwidth]{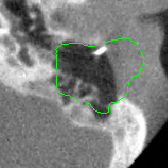}
        \footnotesize{UNET++}
    \end{minipage}
\caption{\textbf{Representative Samples}. Qualitative visualizations of predicted mastoidectomy regions that are highlighted in green using different methods. The \textbf{first row} shows mastoidectomy shape predictions (in green) on preoperative images, while the \textbf{second row} shows predictions on postoperative images.}
\label{fig:comparing_samples}
\end{figure}
\begin{table*}[ht]
\centering
\begin{tabular}{l | lccccccc}
\hline
Methods & Dice$\uparrow$ & IoU$\uparrow$ & Acc$\uparrow$ & Pre$\uparrow$ & Sen$\uparrow$ & Spe$\uparrow$ & HD95$\downarrow$ & ASD$\downarrow$ \\
\hline
UNET++ & 0.328$^{*}$ & 0.203 & 0.889 & 0.563 & 0.244 & 0.977 & 32.367 & 9.349$^*$ \\
R2UNET & 0.484$^{*}$ & 0.325 & 0.914 & 0.848 & 0.348 & 0.992 & 27.948 & 5.273$^*$ \\
R2AttUNET & 0.497$^{*}$ & 0.343 & 0.917 & \textbf{0.891} & 0.361 & \textbf{0.995} & 24.377 & 5.435$^*$ \\
UNETR & 0.698$^{*}$ & 0.542 & 0.931 & 0.821 & 0.624 & 0.983 & 17.612 & 3.890$^*$ \\
Baseline & 0.702  & 0.545 & 0.938 & 0.797 & 0.644 & 0.980 & 17.352 & 4.159$^{*}$ \\
SwinUNETR & 0.712 & 0.557 & 0.939 & 0.806 & 0.655 & 0.981 & 16.930 & 3.971$^{*}$ \\
Proposed Method & \textbf{0.721} & \textbf{0.568} & \textbf{0.941} & 0.826 & \textbf{0.656} & 0.983 & \textbf{16.159} & \textbf{3.638} \\
\hline
\end{tabular}
\caption{\textbf{Quantitative Evaluation}. The proposed method achieves state-of-the-art performance in most metrics compared to widely-used U-Net-based and transformer-based networks. Note: IoU, Acc, Pre, Sen, Spe are the abbreviations of Intersection over Union, Accuracy, Precision, Sensitivity, and Specification, respectively.}
\label{tab:table1}
\end{table*}
Our dataset consists of 751 images with 504 training, 126 validation, and 121 testing samples. In the test set, we manually annotate the ground truth mastoidectomy volume labels for 32 random testing cases. The annotation process, which took approximately 1 hour per case, was labor-intensive due to the heterogeneity between preoperative and postoperative CT scans and the significant variability in the shape and size of the removed mastoid regions. To evaluate the model's ability to segment the mastoidectomy shape within preoperative CT scans, we begin by querying the Mamba-based model with voxel values from the 3D preoperative images and binarizing the predicted probability masks. The results are then used to compare with outputs from the baseline method \cite{zhang2024mmunsupervisedmambabasedmastoidectomy} and the annotated ground truth labels. The results of four mastoidectomy shape predictions are shown in Figure~\ref{fig:ws_output}.
Further representative samples are shown in Figure~\ref{fig:comparing_samples} to demonstrate the promising results of identifying target region by the proposed method when comparing with other state-of-the-art models, such as baseline \cite{zhang2024mmunsupervisedmambabasedmastoidectomy}, SwinUNETR \cite{hatamizadeh2022swinunetrswintransformers}, UNETR \cite{hatamizadeh2021unetrtransformers3dmedical}, Recurrent Residual U-Net (R2UNET) \cite{alom2018recurrentresidualconvolutionalneural}, Recurrent Residual Attention U-Net (R2AttUNET) \cite{r2aunet}, and the vanilla UNET++ \cite{zhou2018unetnestedunetarchitecture}. Note that all comparing models are trained with only $\mathcal{L}_{TD}$ and the best performance checkpoint is used for the evaluation.
Quantitative evaluation is presented in Table~\ref{tab:table1}. The results show that the proposed method outperforms others in most metrics. We obtain the state-of-the-art Dice score of \textbf{0.721} and achieve the lowest HD95 of \textbf{16.159}. 
The asterisk ($^*$) followed by a number indicates that the difference between the proposed and competing methods is statistically significant, as determined by a Wilcoxon test with a \textit{p}-value $\leq$ 0.05 for the Dice and ASD metrics. High precision and specificity indicate that the predicted areas are generally smaller than the ground truth labels, leading to fewer false positives. Sensitivity, on the other hand, measures the model's ability to minimize false negatives, and higher sensitivity means fewer false negatives.
To assess the effectiveness of the proposed 3D T-Distribution Loss in handling noisy labels compared to other commonly used loss functions, we present an ablation study in Table~\ref{tab:table2}. It compares the proposed 3D T-Distribution loss with commonly used loss functions, including Mean Squared Error (MSE), Mean Absolute Error (MAE), Focal Loss (FL), Binary Cross Entropy (BCE), and Cross Entropy (CE), under identical training configurations. 
\begin{table}[t]
    \centering
    \begin{tabular}{p{2cm}cc}
        \hline
        \multirow{2}{*}{\textbf{Loss Function}} & \multicolumn{2}{c}{\textbf{Average Accuracy}} \\
        & Dice$\uparrow$ & HD95$\downarrow$ \\
        \hline
        $\mathcal{L}_{CE}$ & 0.667 $\pm$ 0.082 & 31.163 $\pm$ 8.214 \\
        $\mathcal{L}_{BCE}$ & 0.707 $\pm$ 0.069 & 17.780 $\pm$ 8.057\\
        $\mathcal{L}_{FL}$ & 0.707 $\pm$ 0.066 & 17.124 $\pm$ 7.246 \\
        $\mathcal{L}_{MSE}$ & 0.716 $\pm$ 0.064 & 16.932 $\pm$ 7.812 \\
        $\mathcal{L}_{MAE}$ & 0.719 $\pm$ 0.064 & 16.616 $\pm$ 7.301\\
        $\mathcal{L}_{TD}$ & \textbf{0.721 $\pm$ 0.066} & \textbf{16.159 $\pm$ 6.905} \\
        \hline
    \end{tabular}
    \caption{\textbf{Ablation Study}. Comparing the Dice and HD95 metrics with various widely-used loss functions. The T-Distribution loss function achieved the highest Dice of 0.721 and the lowest HD95 of 16.159.}
    \label{tab:table2}
\end{table}
As shown in Table~\ref{tab:table2}, $\mathcal{L}_{TD}$ is more robust to handle noisy labels and achieves the highest Dice score and the lowest HD95 value.
\section{Conclusion}
\label{sec:conclusion}
In this study, we demonstrate for the first time that mastoidectomy shape prediction and segmentation can be effectively achieved through weakly-supervised training without any labor-intensive human labeling or annotation. By incorporating a Mamba-based neural network architecture and 3D T-Distribution loss function, we have improved the segmentation accuracy and achieved state-of-the-art performance compared to several popular methods. As comprehensive experiments shown in Section~\ref{sec:results}, the proposed framework shows superior performance in predicting mastoidectomy shapes. The robustness of the 3D T-Distribution loss function shows greater tolerance to noise and outliers than commonly used traditional loss functions. In summary, our work offers a robust and promising solution for the weakly-supervised medical image segmentation community. One approach to automatically generate weak labels is by utilizing pre-trained models or traditional atlas-based registration/segmentation methods to reduce or eliminate the need for manual labeling. In the future, it is possible to extend and apply this framework to other types of 3D medical imaging datasets, potentially benefiting a wide range of preoperative planning tasks. Further exploration of the 3D T-Distribution loss in other medical imaging segmentation applications may uncover its broader applicability.

\acknowledgments{This research is supported by grants R01DC014037 and R01DC008408 from the NIDCD. The content is solely the responsibility of the authors and does not necessarily reflect the views of this institute.}

\bibliography{citations}
\bibliographystyle{spiebib}
\end{document}